\documentclass[a4paper,fleqn]{cas-sc}

\makeatletter
\@ifpackageloaded{endfloat}{
  \endfloatdisable 
  \PackageWarning{myfix}{endfloat package detected and disabled!}
}{}
\makeatother

\usepackage[numbers,sort&compress]{natbib}
\usepackage[switch]{lineno}

\usepackage{algorithm}
\usepackage{algorithmic}
\usepackage{amsfonts}
\usepackage{multirow}
\usepackage{makecell}
\usepackage{array}
\usepackage{tabularx}

\DeclareMathOperator*{\st}{s.t.}

\def\tsc#1{\csdef{#1}{\textsc{\lowercase{#1}}\xspace}}
\tsc{WGM}
\tsc{QE}
\tsc{EP}
\tsc{PMS}
\tsc{BEC}
\tsc{DE}

\begin{document}
\setlength{\abovedisplayskip}{8pt}
\setlength{\belowdisplayskip}{8pt}
\let\WriteBookmarks\relax
\def\floatpagepagefraction{1}
\def\textpagefraction{.001}
\shorttitle{Defending against transferable adversarial attacks using MAPE}
\shortauthors{Xinlei Liu et~al.}

\title [mode = title]{MAPE: Defending Against Transferable Adversarial Attacks Using Multi-Source Adversarial Perturbations Elimination}

\author[1]{Xinlei Liu}[]
\cormark[1]

\author[1]{Jichao Xie}[]
\cormark[1]

\author[1]{Tao Hu}[]
\cormark[2]
\ead{hutaondsc@163.com}

\author[1,2]{Peng Yi}[]
\cormark[2]
\ead{yipengndsc@163.com}

\author[1]{Yuxiang Hu}[]

\author[1]{Shumin Huo}[]

\author[1]{Zhen Zhang}[]

\affiliation[1]{organization={Information Engineering University},
                city={Zhengzhou},
                postcode={450002},
                country={China}}

\affiliation[2]{organization={Key Laboratory of Cyberspace Security, Ministry of Education},
                city={Zhengzhou},
                postcode={450002},
                country={China}}

\cortext[cor1]{These authors contributed equally to this work.}
\cortext[cor2]{Corresponding author}

\begin{abstract}
Neural networks are vulnerable to meticulously crafted adversarial examples, leading to high-confidence misclassifications in image classification tasks. Due to their consistency with regular input patterns and the absence of reliance on the target model and its output information, transferable adversarial attacks exhibit a notably high stealthiness and detection difficulty, making them a significant focus of defense. In this work, we propose a deep learning defense known as \textbf{multi-source adversarial perturbations elimination} (MAPE) to counter diverse transferable attacks. MAPE comprises the \textbf{single-source adversarial perturbation elimination} (SAPE) mechanism and the \textbf{pre-trained models probabilistic scheduling algorithm} (PPSA). SAPE utilizes a thoughtfully designed channel-attention U-Net as the defense model and employs adversarial examples generated by a pre-trained model (e.g., ResNet) for its training, thereby enabling the elimination of known adversarial perturbations. PPSA introduces model difference quantification and negative momentum to strategically schedule multiple pre-trained models, thereby maximizing the differences among adversarial examples during the defense model's training and enhancing its robustness in eliminating adversarial perturbations. MAPE effectively eliminates adversarial perturbations in various adversarial examples, providing a robust defense against attacks from different substitute models. In a black-box attack scenario utilizing ResNet-34 as the target model, our approach achieves average defense rates of over 95.1\% on CIFAR-10 and over 71.5\% on Mini-ImageNet, demonstrating state-of-the-art performance.
\end{abstract}


\maketitle

\section{Introduction}
\begin{figure}[t]
  \centering
  \includegraphics[width=0.45\linewidth]{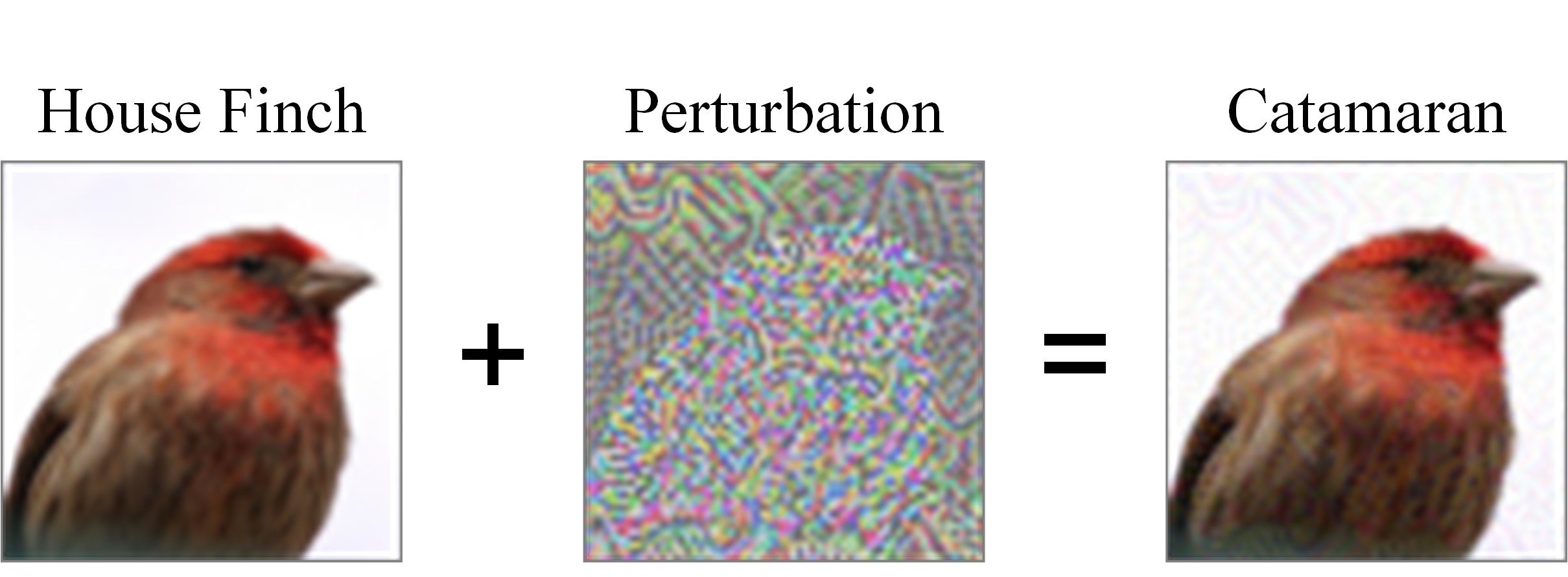}
  \caption{Clean example and adversarial example. When the adversarial perturbations are added to a house finch, it is misclassified as a catamaran by the classification model.}
  \label{Fig1}
\end{figure}

The convolutional neural network (CNN) is a deep and feedforward neural network that incorporates convolution operations, which has been used widely in diverse visual tasks \cite{LeC15}, including image recognition \cite{HeDee16}, object detection \cite{Guo21}, and semantic segmentation \cite{Sia18}. However, recent research has shown that there exist adversarial examples \cite{Zho19,Akh21,Gao24, Qia24} that do not affect human judgment but can perplex network models. For instance, when a classification model correctly identifies a house finch in Figure~\ref{Fig1}, and then a meticulously designed adversarial perturbation is introduced, the model misclassifies it as a catamaran. However, from a human perspective, there is not apparent difference in cognition between the example before and after the inclusion of the adversarial perturbation. Adversarial examples can lead to potential security vulnerabilities, thus affecting the reliability and stability of image classification systems \cite{Kur17,Cro20}. Therefore, defending against adversarial attacks has become one of the important challenges in protecting deep learning models and ensuring system security.

Transferable adversarial attack \cite{XieImp19,Lin20,Wan21,Zhu24, Zhi24} is a classic black-box attack, which refers to generating adversarial examples on a substitute model and then using them to deceive the target model. In comparison to another typical black-box attack method, query attacks \cite{Bha18,LiQue20, Shi23}, transferable attacks are more similar to conventional input patterns. As they do not require sending a large number of query examples to the target model, and are entirely independent of the target model and its output information. Due to their higher stealthiness and greater difficulty in detection, along with more relaxed implementation conditions, transferable attacks have become one of the most prevalent adversarial attack methods currently. Consequently, we will designate it as the defensive target of this study. Differing from adversarial training \cite{Mad18,Zha_The19} aimed at enhancing the robustness of the target model, input transformations \cite{Ozd18,Zho19,Akh21} defend against adversarial attacks by applying random transformations to disrupt adversarial perturbations, or by denoising them to eliminate adversarial perturbations. In defending against transferable attacks that are more aligned with real-world scenarios, input transformations demonstrate a more outstanding defense effectiveness \cite{Raf19,Pan20}.

In input transformation, random transformation methods \cite{Bah19,Li23} disrupt the overall structural adversarial perturbation by randomly rotating, scaling, and translating, enabling the target model to correctly classify adversarial examples. However, random transformations also obscure the original data distribution in the input examples, inevitably leading to a significant decrease in its classification accuracy. Denoising methods \cite{Lia18,XieFea19} eliminate adversarial perturbations from input examples by introducing denoising blocks in the classification model or deploying denoisers externally to the model, exhibiting strong specificity and mediocre generalizability. Specifically, the defense effectiveness of denoising methods is stronger when the substitute model is structurally similar to the target model, while it significantly decreases when there is a large difference in structure between the substitute model and the target model.

In this article, we propose the \textit{multi-source adversarial perturbations elimination} (MAPE) to assist target models in defending against diverse transferable attacks. At a low level, a channel-attention U-Net (CAU-Net) is utilized as the defense model, reconstructing the adversarial examples by eliminating the perturbations within them. Subsequently, the defense model is trained by computing the label losses between reconstructed examples and clean examples. As the adversarial examples originate from a single classification model, we refer to this low-level mechanism as \textit{single-source adversarial perturbation elimination} (SAPE). At a high level, we introduce several distinct pre-trained models and propose the \textit{pre-trained models probabilistic scheduling algorithm} (PPSA). Based on the pre-trained model’s output scores and scheduling records, we define two key components in PPSA: model difference probability and negative momentum probability. The former represents an intrinsic characteristic of the model in the scheduling process, while the latter serves as a regularization factor to adjust the usage of models. PPSA effectively combines these two probabilities to maximize the differences between adjacent scheduled pre-trained models. By integrating SAPE and PPSA, MAPE enhances the defense model's robustness and generalization in eliminating adversarial perturbations. Compared to previous defense strategies, MAPE exhibits superior effectiveness in countering adversarial attacks. These have been verified in Section~\ref{Sec4} and Section~\ref{Sec5}. The code is available at \text{\url{https://anonymous.4open.science/r/MAPE-EB7F}}.

We summarize the main contributions as follows:

\begin{itemize}
\item We propose a deep learning defense known as \textit{ multisource adversarial perturbation elimination} (MAPE), which utilizes a CAU-Net as the defense model and is capable of eliminating adversarial perturbations in various adversarial examples.
\item To the best of our knowledge, we are the first to introduce \textit{negative momentum} as a regularization factor for dynamically adjusting the usage of certain elements, as well as to \textit{quantify the model difference} based on the output scores on the same dataset.
\item The evaluation demonstrates that MAPE exhibits strong generalization capability and cross-model defense characteristics, effectively countering transferable adversarial attacks from various substitute models in a black-box attack environment.
\end{itemize}

\section{Related Work}
We study related work from three perspectives: adversarial examples, attack methods for generating transferable adversarial examples, and defense methods for resisting this adversarial examples.

\subsection{Adversarial Examples}
The generation of adversarial examples can be represented as a constrained optimization problem. Let $\mathcal{C}(\cdot)$ be the pre-trained classification model such that $\mathcal{C}(\boldsymbol{x})$ : $\boldsymbol{x} \rightarrow \boldsymbol{\ell}$, where $\boldsymbol{x} \in \mathbb{R}^m$ is a clean example and $\boldsymbol{\ell} \in \mathbb{Z}^+$ is the output of the model. Let $\mathcal{A}(\cdot)$ be the attack method used by the attacker, denoted as $\mathcal{A}(\boldsymbol{\theta}, \boldsymbol{x}) \rightarrow \boldsymbol{\rho}$, where $\boldsymbol{\theta}$ is the parameter of the model and $\boldsymbol{\rho} \in \mathbb{R}^m$ is the adversarial perturbation. To ensure that the semantic information in natural examples used for human recognition is not compromised, the generated adversarial perturbation $\boldsymbol{\rho}$ is often bounded by a norm. For example, constraining the perturbation $\boldsymbol{\rho}$ within $\|\boldsymbol{\rho}\|_p < \epsilon$, where $\|\boldsymbol{\rho}\|_p$ denotes the $L_p$ norm and $\epsilon$ is the adversarial perturbation budget. The adversarial example $\boldsymbol{\bar{x}} \in \mathbb{R}^m$ is obtained by adding the adversarial perturbation $\boldsymbol{\rho}$ to the natural example $\boldsymbol{x}$, represented as $\boldsymbol{x} + \boldsymbol{\rho} \rightarrow \boldsymbol{\bar{x}}$. The generation problem of adversarial examples is essentially the problem of solving adversarial perturbations, which can be represented by the following constrained optimization process:
\begin{equation}
  \ \ \ \ \ \ \ \ \ \ \ \ \ \ \ \ \ \ \ \ \ \ \ \ \ \ \ \ \ \ \ \ \ \ \ \ \ \ \ \ \ \ \ \ \ \ \ \ \ \ \ \ \underset{\boldsymbol{\rho}}{{\arg\max}}\ \mathcal{L}\left(\boldsymbol{\theta}, \boldsymbol{x} + \boldsymbol{\rho}, \boldsymbol{\ell}\right)\ \st \|\boldsymbol{\rho}\|_p < \epsilon \ .
  \label{Equ1}
\end{equation}

In Equation~\ref{Equ1}, $\mathcal{L}\left(\boldsymbol{\theta}, \boldsymbol{x} + \boldsymbol{\rho}, \boldsymbol{\ell}\right)$ represents the loss of the model with parameter $\boldsymbol{\theta}$ regarding adversarial example $\boldsymbol{x} + \boldsymbol{\rho}$ and label $\boldsymbol{\ell}$, typically computed using the cross-entropy loss function. Therefore, the generation of adversarial examples can be summarized as finding limited adversarial perturbations that maximize the model's loss. Adversarial examples are typically generated through gradient-based methods \cite{Goo15,Kur17} when the model's parameters and defense strategies are known to the attacker. In cases where the model's parameters and defense strategies are unseen to the attacker, adversarial examples are usually generated from substitute models based on their transferability \cite{Cro20,XieImp19,Wan21}.

\subsection{Attack Methods}
Transferable attacks are built on the transferability of adversarial examples \cite{Guo20}. Therefore, the simplest transferable attacks use white-box attack methods as their means of attack, with the difference from regular white-box attacks being that substitute models are treated as target models. Representative methods include the fast gradient sign method (FGSM) \cite{Goo15}, the basic iterative method (BIM) \cite{Kur17}, and the projected gradient descent (PGD) \cite{Mad18}. The FGSM is a one-step gradient-based method that computes norm-bounded perturbations, while BIM and PGD seek better solutions by optimizing the gradient direction through multiple iterations \cite{Akh21}. The core concept of the three methods mentioned above is to perform gradient ascent on the loss surface of the model to deceive it, which also forms the basis of many adversarial attacks.

However, some stronger transferable attacks improve the transferability of adversarial examples by integrating attack techniques, transforming images, optimizing gradient variances, among other strategies. Diverse inputs iterative FGSM (DIM) \cite{XieImp19} applies random image transformations, diversifying the input information for each iteration to enhance the transferability of adversarial examples. Built upon momentum iterative FGSM (MI-FGSM) \cite{Don18} and Nesterov iterative FGSM (NI-FGSM) \cite{Lin20}, respectively, variance tuning MI-FGSM (VMIM) and variance tuning NI-FGSM (VNIM) \cite{Wan21} adjust the current gradient by using the gradient variance from the previous iteration to optimize the gradient direction and escape local optima. Additionally, there are also attack methods that aim to enhance the transferability of adversarial examples by integrating gradients from multiple iterations or multiple models, such as large geometric vicinity (LGV) \cite{Gub22}, transferable adversarial attack based on integrated gradients (TAIG) \cite{Hua22} and adaptive model ensemble adversarial attack (AdaEA) \cite{Che23}. 

\subsection{Defense Methods}

Adversarial defense methods are generally divided into two main classes, including adversarial training and input transformation. Adversarial training \cite{Mad18} is the data augmentation technique that enhances the robustness of the target model by adding adversarial examples to the training data. TRADES \cite{Zha_The19} decomposes the robust error of adversarial examples into the sum of natural error and boundary error, which acts as its design principle in defending against adversarial attacks.

Input transformation methods aim to eliminate the attack nature of adversarial examples, thereby reducing the recognition difficulty for the target model. They are the main force in defending against black-box attacks. These methods can be categorized into random transformation methods and denoising methods. In random transformation methods, total variance minimization (TVM) \cite{Guo18} randomly selects a small group of pixels and reconstructs the "simplest" image that does not include adversarial perturbations. Pixel deflection \cite{Pra18} corrupts adversarial perturbations by redistributing the pixel values and applying adaptive soft-thresholding in the wavelet domain. Mixup inference \cite{Pan20} overlays input examples randomly with other clean examples to reduce the adversarial nature of the input examples. In denoising methods, JPEG compression \cite{Dzi16} removes certain high-frequency components and image details through discrete cosine transformation and quantization, thereby enabling defense against adversarial examples with low perturbation budgets.   Similarly, Gaussian blurring (smoothing) \cite{GauWan21} convolves a Gaussian kernel with adversarial examples, blurring image details to disrupt the adversarial perturbations present in the adversarial examples. Feature denoising method \cite{XieFea19} adds denoising blocks in the classification model and combines them with adversarial training to enhance the model's adversarial robustness. High-level representation guided denoiser (HGD) \cite{Lia18} revises the loss function to pull adversarial examples back to the original clean distribution for improving their classification accuracy. Learning defense transformation (LDT) \cite{Li23} employs parameterizing the affine transformations and the boundary information of neural network as a defense mechanism against adversarial attacks.

\section{Methodology}

Random transformation methods, such as TVM and pixel deflection, do not depend on the target model, resulting in similar defense effectiveness against both known and unknown types of adversarial attacks, although neither is particularly high. Deep learning defense methods, such as HGD and LDT, are closely coupled with the target model. Due to the differences between the target model and the substitute model, their effectiveness in defending against unknown types of adversarial attacks is significantly diminished.

To improve defense effectiveness against unknown types of adversarial attacks, we propose a deep learning defense known as multi-source adversarial perturbations elimination (MAPE). MAPE primarily consists of the single-source adversarial perturbation elimination (SAPE) mechanism and the pre-trained models probabilistic scheduling algorithm (PPSA). SAPE serves as the foundational method for MAPE, aiming to enable the defense model to eliminate known adversarial perturbations. PPSA acts as the organizational framework of MAPE, focusing on achieving the ability to eliminate unknown types of adversarial perturbations and improving its robustness.

\subsection{Single-Source Adversarial Perturbation Elimination}

\begin{figure*}[t]
  \centering
  \includegraphics[width=0.92\linewidth]{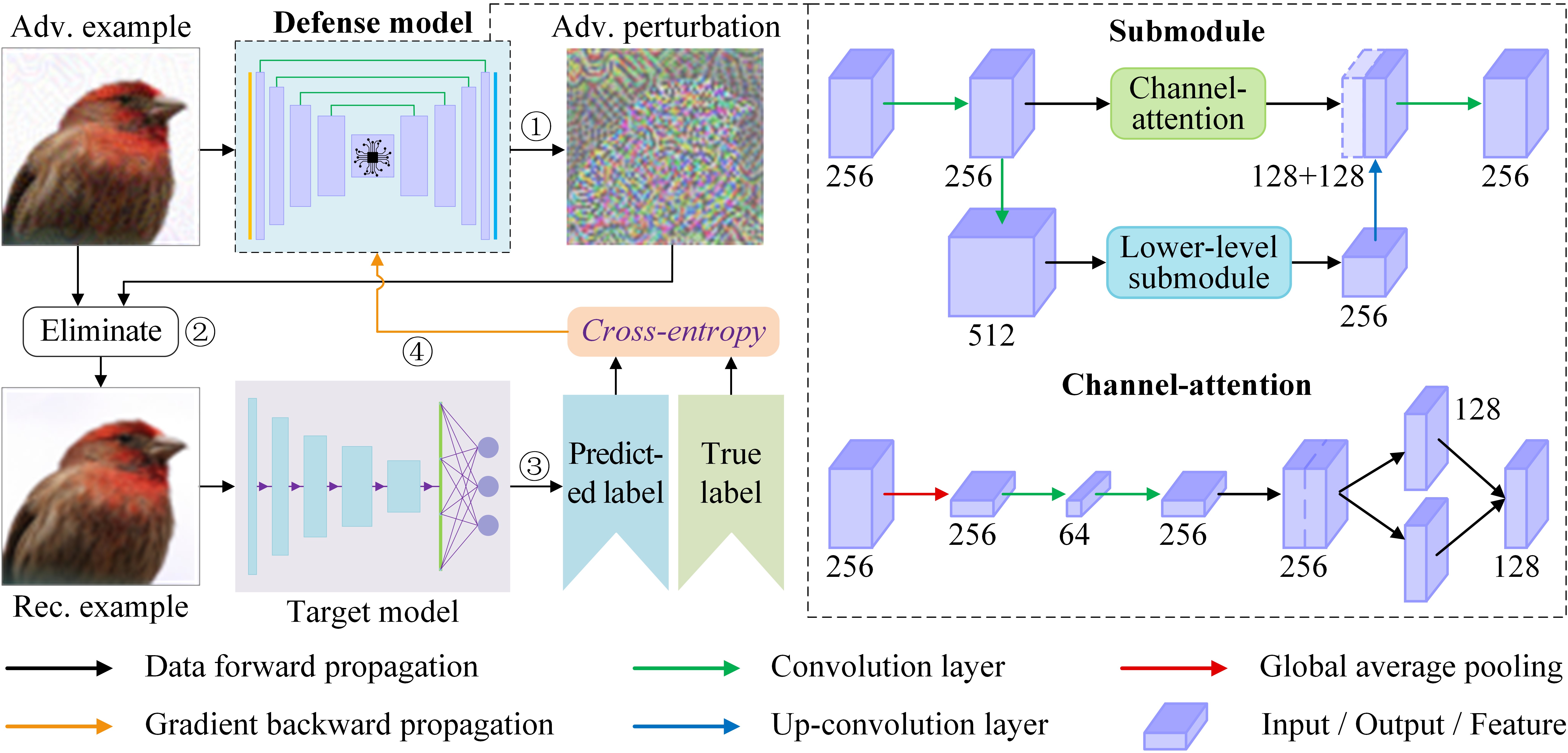}
  \caption{Single-source adversarial perturbation elimination (SAPE) mechanism.}
  \label{Fig2}
\end{figure*}

As shown in Figure~\ref{Fig2}, SAPE consists mainly of a target model $\mathcal{C}(\cdot)$ with the parameter $\boldsymbol{\theta}$ and a defense model $\mathcal{E}(\cdot)$ with the parameter $\boldsymbol{\zeta}$. The target model $\mathcal{C}(\cdot)$ is the commonly used classification model such as ResNet \cite{HeDee16}, GoogLeNet \cite{Chr15}, MobileNet \cite{Mar18}, etc. The defense model $\mathcal{E}(\cdot)$ is typically a neural network model for image-to-image generation. In this paper, we integrate a U-Net \cite{Ron15} with channel-attention mechanism \cite{Hu18} as a defense model, referred to as CAU-Net. Compared to generative adversarial networks (GANs) \cite{Goo14} and diffusion models \cite{Ho20}, U-Net requires less computational cost and is easier to train. Additionally, the primary comparison methods, HGD and LDT, also utilize U-Net or its variants as defense models; therefore, selecting U-Net enhances the credibility of the experimental results. We also optimize its structure to enhance its capability in extracting adversarial perturbations. The optimized CAU-Net is constructed with multiple nested submodules. Except for the lowest-level submodules, each submodule nests a lower-level submodule and contains a channel-attention mechanism layer similar to a residual connection. This work utilizes a CAU-Net with five submodules. The right side of Figure~\ref{Fig2} illustrates the third submodule along with its specific components. The number of both input and output data channels is 256.

Deployed outside the target model, the defense model $\mathcal{E}(\cdot)$ is responsible for extracting and eliminating the adversarial perturbation $\boldsymbol{\hat{\rho}}$ from the adversarial example $\boldsymbol{\bar{x}}$, playing a role similar to that of antivirus software. Note that the adversarial perturbation $\boldsymbol{\hat{\rho}}$ extracted from adversarial examples by the defense model is not equivalent to the adversarial perturbation $\boldsymbol{\rho}$ added by an attacker to clean examples. It is desirable to enhance the similarity of the data distributions between the two, which is the goal pursued by the defense model. 

The training method of SAPE is as follows. Firstly, the adversarial example generated by the target model is input into the defense model to extract the adversarial perturbation, which is then removed to obtain the reconstructed example $\boldsymbol{\hat{x}}$. Subsequently, the reconstructed example is fed into the target model to obtain the probability distribution of predicted labels. Finally, the cross-entropy is computed between the probability distribution of the predicted label and that of the true label, after which the defense model's weights are updated with the back-propagation algorithm. The optimization objective of SAPE can be expressed by Equation~\ref{Equ2}.
\begin{equation}
  \ \ \ \ \ \ \ \ \ \ \ \ \ \ \ \ \ \ \ \ \ \ \ \ \ \ \ \ \ \ \ \ \ \ \ \ \ \ \ \ \ \ \ \ \ \ \ \ \ \ \ \ \underset{\boldsymbol{\zeta}}{{\arg\min}}\ \mathcal{L}\left[\boldsymbol{\theta}, \boldsymbol{x} + \boldsymbol{\rho} - \mathcal{E}(\boldsymbol{x} + \boldsymbol{\rho}), \boldsymbol{\ell}\right]\ .
  \label{Equ2}
\end{equation}

SAPE aims to equip the defense model with the capability to eliminate known adversarial perturbations. This capability serves as the foundation for MAPE. Algorithm~\ref{Alg1} summarizes the detailed training method of SAPE.

\begin{algorithm}[t]
	\caption{Detailed training method of SAPE.}
	\label{Alg1}
	\begin{algorithmic}[1]
		\REQUIRE A defense model $\mathcal{E}$ with the parameter $\boldsymbol{\zeta}$, the target classification model (e.g., ResNet) with the parameter $\boldsymbol{\theta}$, clean examples $\boldsymbol{x}$, attack method $\mathcal{A}$, learning rate $\eta$ and weight decay $\lambda$
		\ENSURE The well-trained defense model $\mathcal{E}$
		\STATE Initiate the parameter $\boldsymbol{\zeta}$ of the defense model $\mathcal{E}$\ ;
        \STATE Freeze the parameters $\boldsymbol{\theta}$ of the target model;
		\WHILE{\text{ not converged }}
		\STATE Generate the adversarial example $\boldsymbol{\bar{x}} = \boldsymbol{x} + \mathcal{A}(\boldsymbol{\theta}, \boldsymbol{x})$\ ;
        \STATE Perform stratified sampling from clean examples and adversarial examples to create a mixed example\\
        $\boldsymbol{\ddot{x}}\leftarrow[\boldsymbol{x},\boldsymbol{\bar{x}}]$\ ;
		\STATE Extract and eliminate the adversarial perturbation $\boldsymbol{\hat{x}} = \boldsymbol{\ddot{x}} - \mathcal{E}(\boldsymbol{\ddot{x}})$\ ;
		\STATE Compute the cross-entropy loss $l = \mathcal{L}(\boldsymbol{\theta}, \boldsymbol{\hat{x}}, \boldsymbol{\ell})$\ ;
		\STATE Update the parameter $\boldsymbol{\zeta} \leftarrow \boldsymbol{\zeta}-\eta \{ \nabla_{\boldsymbol{\zeta}} l + \lambda \boldsymbol{\zeta} \}$\ .
		\ENDWHILE
	\end{algorithmic}
\end{algorithm}

\subsection{Pre-Trained Models Probabilistic Scheduling Algorithm}

During training SAPE, the adversarial examples used are solely derived from the target model. When the attacker's substitute model differs from the protected target model, the defense effectiveness of SAPE significantly decreases. In practical applications, it is quite common for the substitute model to differ from the target model.

Therefore, utilizing multiple pre-trained models and enhancing the difference among these models can diversify the adversarial examples used for training, thereby endowing the defense model with higher generalization capability and defensive performance. And the impact of model quantity on defensive performance depends on the differences between the newly added models and the previous models. Furthermore, to fully leverage the difference among these pre-trained models for training the defense model, we propose PPSA to strategically schedule them. PPSA refers to selecting a different pre-trained model based on the scheduling probability after training on current mini-batch. The newly selected pre-trained model will be used to generate adversarial examples of the next mini-batch. If the pre-trained model $C_i(i=1,2,\cdots,N)$ is used to generate adversarial examples in the $k\text{-th}$ mini-batch, then the probability of the pre-trained model $C_j(j=1,2,\cdots,N,j\neq i)$ being selected in the $(k+1)\text{-th}$ mini-batch can be expressed as
\begin{equation}
  \ \ \ \ \ \ \ \ \ \ \ \ \ \ \ \ \ \ \ \ \ \ \ \ \ \ \ \ \ \ \ \ \ \ \ \ \ \ \ \ \ \ \ \ \ \ \ \ \ \ \ \ \ \ \ \ \ \ \ \ \ \ P^{j}_{k}=h\left\{P^{(i,j)}_{\text{diff}}\circ P^{(k,j)}_{\text{neg}}\right\}\ .
  \label{Equ3}
\end{equation}
In Equation~\ref{Equ3}, $h\{\cdot\}$ is the probability normalization transformation. For the variable $I_s(s=1,2,\cdots,N)$,
\begin{equation}
  \ \ \ \ \ \ \ \ \ \ \ \ \ \ \ \ \ \ \ \ \ \ \ \ \ \ \ \ \ \ \ \ \ \ \ \ \ \ \ \ \ \ \ \ \ \ \ \ \ \ \ \ \ \ \ \ \ \ \ \ \ \ \ \ \ \ h\left\{I_s\right\}=\frac{I_s}{\sum^{N}_{t=1}{I_t}}\ .
  \label{Equ4}
\end{equation}
$P^{(i,j)}_{\text{diff}}$ represents the model difference probability between the pre-trained models $C_i$ and $C_j$, $P^{(k,j)}_{\text{neg}}$ represents the negative momentum probability of $C_j$ at the $k\text{-th}$ mini-batch, and "$\circ$" denotes their Hadamard product. Below, we will discuss these two probability distributions separately.

Pre-trained models with greater differences are usually selected to generate adversarial examples during the training of adjacent mini-batches, as this can help the defense model adapt to diverse adversarial inputs and boosts its robustness against adversarial attacks. However, it is difficult to quantify the differences between different models by comparing their structures and parameters. We think that a model's output scores reflect its intrinsic characteristics to some extent; a larger difference in output scores indicates a greater difference between two models. Therefore, we propose a model differences calculating method based on output score. Suppose that pre-trained models have output scores $E_i(i=1,2,\cdots,N)$ on the same dataset, then the difference $W_{i,j}$ between models $C_i$ and $C_j$ can be expressed using $L_{1}$ norm Wasserstein distance as:
\begin{equation}
   \ \ \ \ \ \ \ \ \ \ \ \ \ \ \ \ \ \ \ \ \ \ \ \ \ \ \ \ \ \ \ \ \ \ \ \ \ \ \ \ \ \ \ \ \ \ \ \ \ \ \ W_{i,j}=\inf_{\gamma\in\Gamma\left(E_i,E_j\right)}\int_{\mathbb{R}\times\mathbb{R}}|u-v|\mathrm{d}\gamma(u, v)\ .
  \label{Equ5}
\end{equation}
In Equation~\ref{Equ5}, $\Gamma(E_i,E_j)$ is the set of all joint distributions whose marginal distributions are $E_i$ and $E_j$, respectively. $\gamma$ represents a joint distribution that describes how to "transport" or "transfer" probability mass between the two distributions. $W_{i,j}$ has good mathematical properties, such as non-negativity $\left(W_{i,j}\geq 0\right)$ and symmetry $\left(W_{i,j}=W_{j,i}\right)$. Notably, the symmetry property is not present in cross-entropy and KL divergence. A higher value of $W_{i,j}$ indicates a greater difference between models $C_i$ and $C_j$. After calculating the differences of the pre-trained models, the model difference probability can be expressed as
\begin{equation}
   \ \ \ \ \ \ \ \ \ \ \ \ \ \ \ \ \ \ \ \ \ \ \ \ \ \ \ \ \ \ \ \ \ \ \ \ \ \ \ \ \ \ \ \ \ \ \ \ \ \ \ \ \ \ \ \ \ \ \ \ P^{(i,j)}_{\text{diff}}=h\left\{\ln\left(W_{i,j}+1\right)\right\}\ .
  \label{Equ6}
\end{equation}

The model difference probability is a static attribute of the pre-trained models, remaining unchanged throughout the entire training period. This leads to a stable scheduling ratio among the various pre-trained models as training iterations increase, which is detrimental to the defense model's generalization. To address this issue, we propose the negative momentum probability as a regularization factor to dynamically adjust the model difference probability, expressed as
\begin{equation}
   \ \ \ \ \ \ \ \ \ \ \ \ \ \ \ \ \ \ \ \ \ \ \ \ \ \ \ \ \ \ \ \ \ \ \ \ \ \ \ \ \ \ \ \ \ \ \ \ \ \ \ \ \ \ \ \ \ \ \ P^{(k,j)}_{\text{neg}}=h\left\{1-h\left\{M_{k,j}\right\}\right\}\ .
  \label{Equ7}
\end{equation}
In Equation~\ref{Equ7}, $M_{k,j}$ represents the total number of times model $C_j$ has been selected up to the $k\text{-th}$ mini-batch. It can be observed that $P^{(k,j)}_{\text{neg}}$ is negatively correlated with $M_{k,j}$, meaning that for pre-trained models that are frequently utilized, $P^{(k,j)}_{\text{neg}}$ will decrease the their scheduling probability;  conversely, for models that are rarely used, $P^{(k,j)}_{\text{neg}}$ will increase their scheduling probability. Contrary to the effect of traditional "momentum," $P^{(k,j)}_{\text{neg}}$ suppresses the excessive use of high-difference-probability models during training, thereby enhancing the defense model's generalization toward other low-difference-probability models. Hence, $P^{(k,j)}_{\text{neg}}$ is referred to as "negative momentum" probability.

\begin{algorithm}[t]
	\caption{Detailed scheduling method of PPSA.}
	\label{Alg2}
	\begin{algorithmic}[1]
		\REQUIRE The pre-trained model $C_i(i=1,2,\cdots,N)$ used in current mini-batch, output scores $E$ of all pre-trained models on the same dataset, and their total scheduling times $M_k$ up to the $k\text{-th}$ mini-batch.
		\ENSURE The scheduled pre-trained model $C_r$ in $(k+1)\text{-th}$ mini-batch
        \STATE Compute the model difference $W_{i,j}$ between the current pre-trained model $C_i$ and the remaining pre-trained models $C_j(j=1,2,\cdots,N,j\neq i)$ based on $L_1$ norm Wasserstein distance;
        \STATE Get the model difference probability $P^{(i,j)}_{\text{diff}}$ according to the model difference $W_{i,j}$\ ;
        \STATE Get the negative momentum probability $P^{(k,j)}_{\text{neg}}$ according to the total scheduling times $M_k$\ ;
        \STATE Generate the scheduling probability $P^{j}_{k}$ of the pre-trained model $C_j$ by combining $P^{(i,j)}_{\text{diff}}$ and $P^{(k,j)}_{\text{neg}}$\ ;
		\STATE Determine the pre-trained model $C_r$ based on the scheduling probability $P^{j}_{k}$\ ;
        \STATE Update the total scheduling times of the pre-trained model $C_r$: $M_{k+1,r}=M_{k,r}+1$\ .
	\end{algorithmic}
\end{algorithm}

PPSA combines model difference probability and negative momentum probability, dynamically scheduling different pre-trained models to generate diverse adversarial examples based on the current state and historical records. This approach helps improve the robustness of the defense model's capability in eliminating unknown types of adversarial perturbations. Algorithm~\ref{Alg2} summarizes the detailed scheduling method of the PPSA.

\subsection{Multi-Source Adversarial Perturbations Elimination}

\begin{figure}[t]
  \centering
  \includegraphics[width=0.7\linewidth]{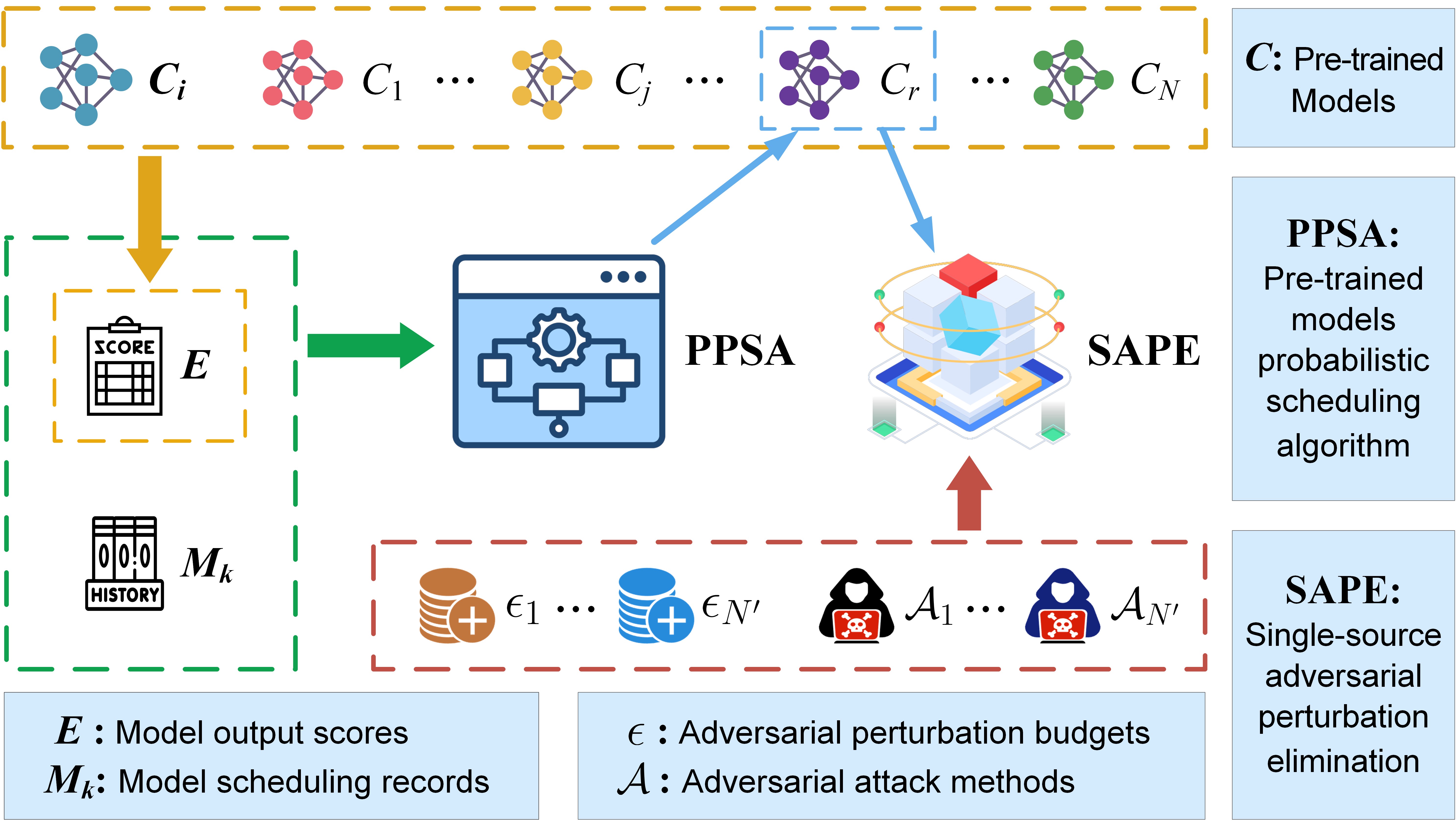}
  \caption{Deep learning defense known as multi-source adversarial perturbations elimination (MAPE).}
  \label{Fig3}
\end{figure}

As shown in Figure~\ref{Fig3}, MAPE consists primarily of PPSA and SAPE, with the training method outlined as follows: First, we calculate the output score between the current pre-trained model and the remaining pre-trained model. Then, utilize PPSA to select one of the remaining pre-trained models based on the model output scores and the model scheduling records. Finally, this scheduled pre-trained model is employed as the target model in SAPE for training the next mini-batch. Notably, to enhance the diversity of the training process, we use "mini-batch" as the switching cycle for pre-trained models instead of "epoch." The model scheduling records are preserved throughout the entire training cycle rather than being reset at the start of a new "epoch." Additionally, compared to SAPE, MAPE incorporates random adversarial perturbation budgets and random attack methods, which contribute to more generalized training for the defense model. Algorithm~\ref{Alg3} summarizes the detailed training method of MAPE.

\begin{figure}[t]
  \centering
  \includegraphics[width=0.85\linewidth]{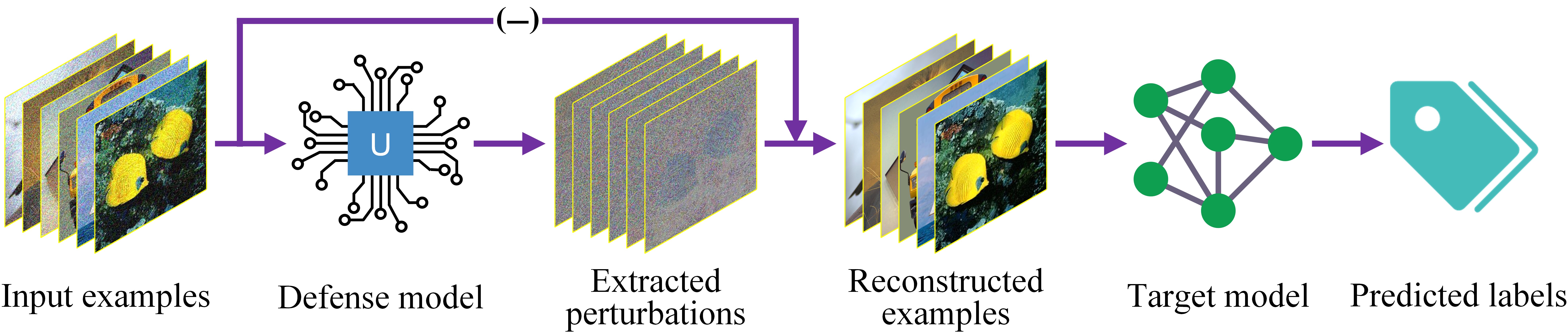}
  \caption{Utilizing MAPE to defend against adversarial attacks.}
  \label{Fig4}
\end{figure}

The process of utilizing MAPE to defend against adversarial attacks is illustrated in Figure~\ref{Fig4}. The defense model is deployed externally to the target model and is responsible for extracting and eliminating implicit adversarial perturbations from input examples, thereby diminishing the effectiveness of adversarial attacks. The reconstructed examples will then be input to the target model for classification.  If the input examples are clean natural images, the extracted adversarial perturbations are meaningless and do not affect normal classification.

\begin{algorithm}[t]
	\caption{Detailed training method of MAPE.}
	\label{Alg3}
	\begin{algorithmic}[1]
		\REQUIRE A defense model $\mathcal{E}$ with the parameter $\boldsymbol{\zeta}$, $N$ pre-trained classification models with the parameters $\boldsymbol{\theta}_n (n=1,2,\dots,N)$, clean examples $\boldsymbol{x}$, attack methods $\mathcal{A}_{n'} (n'=1,2,\dots,N')$, learning rate $\eta$ and weight decay $\lambda$
		\ENSURE Robust defense model $\mathcal{E}$
		\STATE Initiate the parameter $\boldsymbol{\zeta}$ of the defense model $\mathcal{E}$\ ;
        \STATE Freeze the parameters $\boldsymbol{\theta}_n$ of all pre-trained models;
		\WHILE{\text{ not converged }}
        \STATE Schedule a pre-trained model $C_r$ according to \textbf{PPSA};
        \STATE Randomly create a sequence of adversarial perturbation budgets $[\epsilon_1,\epsilon_2,\dots,\epsilon_{N'}] \leftarrow (4/255, 12/255)$\ ;
		\STATE Randomly select an attack method to generate adversarial examples $\boldsymbol{\bar{x}}_{n'} = \boldsymbol{x} + \mathcal{A}_{n'}(\boldsymbol{\theta}_{r_{n'}}, \boldsymbol{x})$\ ;
        \STATE Perform stratified sampling from clean examples and adversarial examples to create a mixed example\\
$\boldsymbol{\ddot{x}}\leftarrow[\boldsymbol{x},\boldsymbol{\bar{x}}_{1},\boldsymbol{\bar{x}}_{2},\dots,\boldsymbol{\bar{x}}_{N'}]$\ ;
		\STATE Execute steps 6, 7, and 8 of \textbf{SAPE}.
		\ENDWHILE
	\end{algorithmic}
\end{algorithm}

\section{Experiments}
\label{Sec4}

\subsection{Experimental Setup}
\label{Sec41}

\textbf{Attackers.} The proposed method focuses on defending against transferable adversarial attacks in image classification. The black-box attack environment represents the most common real-world scenario and is applicable in Subsections~\ref{Sec42}-~\ref{Sec44} and Section~\ref{Sec5}. Additionally, Subsection~\ref{Sec45} discusses the less prevalent scenarios involving gray-box and white-box attacks. In a gray-box attack environment, attackers can acquire training data and model architecture from benign users, creating substitute models that align with the target model and defense architecture but are initialized differently, thereby enabling strong substitute model attacks. In a white-box attack environment, attackers have complete access to the target model, including all information about the model and its defense strategy, enabling precise adaptive attacks. To emphasize the accuracy of the evaluations, all attack methods in this paper belong to the more potent non-targeted adversarial attacks.

\textbf{Datasets.} We use CIFAR-10, CIFAR-100, and Mini-ImageNet as the evaluation datasets for this work. The resolutions of CIFAR-10 and CIFAR-100 remain unchanged, and the resolution of Mini-ImageNet is set to $64\times64$. Mini-ImageNet contains 100 classes, all of which are utilized. For each class, 480 randomly selected images are assigned to the training set, while the remaining 120 images are designated for the test set. In the experiments detailed in this paper, all input example values are constrained within the range of 0 to 1, signifying that the entirety of input data comprises image examples.

\textbf{Classification models.} During the training of MAPE, we employ DenseNet \cite{Hua17}, DPN \cite{Che17}, GoogLeNet \cite{Chr15}, MobileNetV2 \cite{Mar18}, PyramidNet \cite{Han17}, RegNet \cite{Rad20}, ResNet \cite{HeDee16}, ResNeXt \cite{Xie17}, SENet \cite{Hu18}, and WideResNet (WRN) \cite{Zag16} as the pre-trained models for crafting adversarial examples. In the evaluation phase, ResNet \cite{HeDee16}, ResNetV2 \cite{HeIde16}, ShuffleNetV2 \cite{Ma18}, VGG \cite{Kar15}, and Vision Transformer \cite{Dos21} are as substitute models used by attackers and utilized to create adversarial examples for comparing the defense robustness of MAPE and baseline methods. To encompass a wide range of attack sources and achieve more realistic performance evaluations, our selected models include both classic and cutting-edge models, spanning from large-scale to lightweight designs.

\textbf{Baseline defense approaches.} We compare with natural training, adversarial training \cite{Mad18} and the following input transformation defense methods: TVM \cite{Guo18}, feature denoising \cite{XieFea19}, pixel deflection \cite{Pra18}, mixup inference \cite{Pan20}, HGD \cite{Lia18} and LDT \cite{Li23}. Except for natural training, all defense methods incorporate adversarial training to enhance their defense performance. To enhance the persuasiveness of the experiments, we set the target model as the ResNet series, similar to other defense methods. Considering the balance between classification performance and computational cost, we decide to use ResNet-34 as the target model of all defense methods. Feature denoising employs a ResNet-34 model with denoising blocks as its target model.

\textbf{Training details.} During the training process of MAPE, attack methods DIM \cite{XieImp19} and PGD \cite{Mad18} are used to generate adversarial examples. The perturbation budget is within the range of (4/255, 12/255). The step size is set to 2/255, while the number of steps is set to 20. The training data is a mixture of adversarial examples and clean examples. The defense model $\mathcal{E}$ is optimized using Adam. Their initial learning rate $\eta$ and weight decay $\lambda$ are set to 0.01 and 0. The number of training epochs is set to 120, with the learning rate decreasing by a factor of 10 at the 50th, 75th, and 100th epochs. The batch size of the dataset is set to 128. The GPU device used is a NVIDIA Tesla A100 (40GB).

\subsection{Existence of MAPE}
\label{Sec42}

In this section, we conduct a series of controlled experiments on the CIFAR-10, CIFAR-100, and Mini-ImageNet datasets, to validate the effectiveness of MAPE and its components (SAPE and PPSA) in adversarial defense. Each set of controlled experiments uses ResNet-34 as the target model under attack, while both ResNet-34 and unforeseen ShuffleNet-V2-2$\times$ serve as substitute models for launching the attacks. The methods compared included natural training, SAPE, SAPE driven by randomly selected pre-trained models (SAPE \& Random), and SAPE driven by pre-trained models scheduled with PPSA (SAPE \& PPSA, namely MAPE). The evaluation results are presented in Table~\ref{Tab1}.

It can be observed that, compared to natural training, SAPE has significantly improved adversarial defense performance. However, since the defense model is trained solely on adversarial examples generated by the target model ResNet-34, the effectiveness of SAPE in defending against adversarial attacks from the unforeseen substitute model ShuffleNet-V2-2$\times$ significantly decreases. On CIFAR-10, CIFAR-100, and Mini-ImageNet, the average defensive performance of SAPE against ShuffleNet-V2-2$\times$ as a substitute model is lower than that against ResNet-34 as a substitute model by 2.97\%, 5.09\%, and 7.16\%, respectively. This indicates that the defense model trained solely with SAPE exhibits weak generalization capability.

In contrast, when randomly selecting pre-trained models to drive SAPE (SAPE \& Random) for training the defense model, it shows relatively close defense performance against attacks from different substitute models. The corresponding data for the aforementioned metrics are 0.38\%, 1.78\%, and 2.74\%, respectively. When using the PPSA to strategically schedule the pre-trained models to drive SAPE (SAPE \& PPSA, namely MAPE),  the defense model exhibits stronger generalization capabilities, with the aforementioned metrics being 0.21\%, 1.10\%, and 1.69\%, respectively. Additionally, due to the dynamic adjustment of the negative momentum mechanism, its defense performance has also been effectively enhanced. On CIFAR-10, CIFAR-100, and Mini-ImageNet, SAPE \& PPSA (MAPE) outperforms SAPE \& Random in average defense effectiveness by 1.55\%, 1.47\%, and 1.95\%, respectively.  These evaluation results indicate that, compared to SAPE, MAPE can effectively enhance both the defense effectiveness and generalization capability of the defense model.

\begin{table*}[t]
  \caption{Classification accuracy rates (\%) of SAPE and MAPE in defending against adversarial attacks on CIFAR-10, CIFAR-100 and Mini-ImageNet (\textit{higher is better}). ResNet-34 serves as the target model. Simultaneously, it and ShuffleNet-V2-2$\times$ serve as substitute models for launching the attacks. For each attack, we show the most successful defense with bold.} \label{Tab1}
  \centering
  \newcolumntype{C}{>{}X}
  \newcolumntype{L}[1]{>{\centering\arraybackslash}p{#1}}
  \newcolumntype{M}[1]{>{\arraybackslash}p{#1}}
  \scalebox{0.88}{
  \begin{tabularx}{\textwidth}{M{1.5cm}C|L{0.77cm}|*{4}{L{0.77cm}}|*{4}{L{0.77cm}}}
    \toprule
    \multirow{2}{*}{\ Datasets} &\multirow{2}{*}{Defenses} & \multirow{2}{*}{Clean} & \multicolumn{4}{c|}{ResNet-34} & \multicolumn{4}{c}{ShuffleNet-V2-2$\times$}\\[0.4ex]
     & & & FGSM & BIM & DIM & PGD & FGSM & BIM & DIM & PGD\\ 
    \cmidrule{1-11}
    \multirow{4}{*}{CIFAR-10}
    & Natural Training & \textbf{95.82} & 27.60 & \phantom{0}0.11 & \phantom{0}0.05 & \phantom{0}0.02 & 51.24 & \phantom{0}9.90 & \phantom{0}6.23 & \phantom{0}7.16\\
    & SAPE & 95.23 & \textbf{95.28} & 95.14 & \textbf{95.08} & 94.15 & 91.99 & 91.86 & 92.13 & 91.78\\
    & SAPE \& Random & 95.02 & 93.66 & 93.90 & 93.65 & 92.90 & 93.43 & 93.06 & 93.21 & 92.90\\
    & SAPE \& PPSA (MAPE) & 95.37 & 95.22 & \textbf{95.42} & 94.98 & \textbf{94.36} & \textbf{95.28} & \textbf{94.76} & \textbf{94.44} & \textbf{94.66}\\
    \cmidrule{1-11}
    \multirow{4}{*}{CIFAR-100}
     & Natural Training & \textbf{75.30} & 16.43 & \phantom{0}1.28 & \phantom{0}1.03 & \phantom{0}0.88 & 24.42 & \phantom{0}6.67 & \phantom{0}3.98 & \phantom{0}4.47\\
     & SAPE & 74.82 & \textbf{74.48} & 72.75 & 72.34 & \textbf{72.25} & 70.96 & 67.27 & 66.63 & 66.59\\
     & SAPE \& Random & 74.54 & 73.31 & 71.49 & 71.36 & 71.18 & 72.01 & 69.65 & 69.32 & 69.25\\
     & SAPE \& PPSA (MAPE) & 74.99 & 74.23 & \textbf{72.93} & \textbf{72.57} & 72.14 & \textbf{74.65} & \textbf{70.64} & \textbf{70.76} & \textbf{71.40}\\
    \cmidrule{1-11}
    \multirow{4}{*}{\makecell[l]{Mini-\\ ImageNet}}
    & Natural Training & \textbf{76.37} & 11.64 & \phantom{0}0.03 & \phantom{0}0.00 & \phantom{0}0.01 & 17.63 & \phantom{0}1.69 & \phantom{0}1.02 & \phantom{0}1.26\\
     & SAPE & 74.62 & \textbf{72.71} & 71.18 & 70.88 & \textbf{71.29} & 65.95 & 64.78 & 63.03 & 63.64\\
     & SAPE \& Random & 74.02 & 71.08 & 69.92 & 69.76 & 69.92 & 68.83 & 67.82 & 65.52 & 67.57\\
     & SAPE \& PPSA (MAPE) & 74.86 & 72.64 & \textbf{71.45} & \textbf{71.09} & 71.20 & \textbf{69.13} & \textbf{70.28} & \textbf{69.85} & \textbf{70.35}\\
    \bottomrule
  \end{tabularx}
  }
\end{table*}

\subsection{Defending Against Unknown Types of Transferable Attacks}
\label{Sec43}

\begin{table*}[t]
  \caption{Classification accuracy rates (\%) of different methods in defending against unknown types of adversarial attacks on CIFAR-10, CIFAR-100 and Mini-ImageNet (\textit{higher is better}). ResNet-34 serves as the target model under attack, while it, along with ShuffleNet-V2-2$\times$ and ResNet-V2-50, acts as substitute models for launching the attacks. For each attack, we show the most successful defense with bold and the second one with underline.}\label{Tab2}
  \centering
  \newcolumntype{C}{>{}X}
  \newcolumntype{L}[1]{>{\centering\arraybackslash}p{#1}}
  \scalebox{0.85}{
  \begin{tabularx}{1.15\textwidth}{CC|L{0.74cm}|*{4}{L{0.74cm}}|*{4}{L{0.74cm}}|*{4}{L{0.74cm}}}
    \toprule
    \multirow{2}{*}{Dataset} & \multirow{2}{*}{Defenses} & \multirow{2}{*}{Clean} & \multicolumn{4}{c|}{ResNet-34} & \multicolumn{4}{c|}{ShuffleNet-V2-2$\times$} & \multicolumn{4}{c}{ResNet-V2-50}\\[0.4ex]
    &&& FGSM & BIM & UPGD & VNIM & FGSM & BIM & UPGD & VNIM & FGSM & BIM & UPGD & VNIM\\
    \cmidrule{1-15}
    \multirow{9}{*}{CIFAR-10}
    &Nat. Tra. & \textbf{95.82} & 27.60 & \phantom{0}0.11 & \phantom{0}0.07 & \phantom{0}0.04 & 51.24 & \phantom{0}9.90 & \phantom{0}6.18 & \phantom{0}2.93 & 46.33 & \phantom{0}8.78 & \phantom{0}4.97 & \phantom{0}2.44\\
    &Adv. Tra. & 88.48 & 86.94 & 88.25 & 87.36 & 86.42 & 85.42 & 86.17 & 85.62 & 84.86 & 85.64 & 86.23 & 85.69 & 85.22\\
    &TVM & 89.32 & 88.05 & 88.24 & 87.67 & 87.04 & 85.99 & 86.41 & 85.95 & 85.38 & 86.59 & 86.88 & 86.64 & 86.01\\
    &Feat. Den. & 88.96 & 87.47 & 88.35 & 87.88 & 87.12 & 85.53 & 86.54 & 85.79 & 85.42 & 86.05 & 86.94 & 86.40 & 86.39\\
    &Pix. Def. & 90.44 & 89.61 & 91.20 & 90.73 & 91.27 & 87.82 & 89.56 & 89.21 & 89.36 & 88.75 & 90.33 & 90.17 & 90.46\\
    &Mix. Inf. & 95.30 & 91.69 & 94.51 & 93.33 & 93.39 & 88.84 & \underline{93.72} & 91.30 & 91.81 & 90.55 & \underline{94.18} & 92.38 & 92.43\\
    &HGD & 93.49 & 93.50 & 93.94 & 93.96 & \underline{94.15} & 91.86 & 92.07 & 91.91 & \underline{93.22} & 92.21 & 92.04 & 93.31 & \underline{93.69}\\
    &LDT & 95.31 & \underline{95.07} & \underline{95.14} & \textbf{95.17} & 93.69 & \underline{93.29} & 93.23 & \underline{93.19} & 92.13 & \underline{93.58} & 93.85 & \underline{93.76} & 93.15\\
    &MAPE & \underline{95.37} & \textbf{95.22} & \textbf{95.42} & \underline{94.23} & \textbf{94.48} & \textbf{95.28} & \textbf{94.76} & \textbf{94.95} & \textbf{95.18} & \textbf{95.19} & \textbf{95.29} & \textbf{94.97} & \textbf{95.02}\\
    \cmidrule{1-15}
    \multirow{9}{*}{CIFAR-100}   
    &Nat. Tra. & \textbf{75.30} & 16.43 & \phantom{0}1.28 & \phantom{0}0.86 & \phantom{0}0.73 & 24.42 & \phantom{0}6.67 & \phantom{0}4.29 & \phantom{0}2.87 & 29.10 & 17.49 & 13.54 & \phantom{0}7.38\\
    &Adv. Tra. & 63.88 & 63.30 & 63.35 & 62.50 & 62.07 & 59.76 & 60.05 & 59.15 & 58.48 & 61.31 & 61.75 & 61.18 & 60.64\\
    &TVM & 65.43 & 62.53 & 62.42 & 61.01 & 60.84 & 59.00 & 59.13 & 57.90 & 57.41 & 61.91 & 62.53 & 61.59 & 60.94\\
    &Feat. Den. & 62.07 & 59.94 & 60.58 & 58.94 & 58.46 & 56.74 & 57.04 & 55.80 & 55.16 & 59.56 & 59.88 & 59.04 & 59.13\\
    &Pix. Def. & 66.83 & 63.98 & 65.49 & 64.62 & 64.67 & 60.56 & 62.12 & 61.09 & 61.39 & 64.08 & 64.74 & 64.26 & 63.91\\
    &Mix. Inf. & 73.67 & 65.06 & 70.98 & 68.85 & 68.76 & 61.94 & \underline{67.61} & 65.83 & 65.72 & 66.67 & \underline{69.10} & 68.02 & 67.50\\
    &HGD & 69.08 & 71.30 & 69.45 & 68.34 & 69.91 & 68.28 & 65.93 & 65.05 & 66.50 & \underline{72.85} & 69.05 & 68.36 & 70.45\\
    &LDT & 74.78 & \textbf{74.59} & \underline{72.55} & \underline{71.41} & \textbf{72.20} & \underline{71.14} & 67.07 & \underline{66.30} & \underline{67.57} & 72.54 & 68.66 & \underline{69.57} & \underline{70.60}\\
    &MAPE & \underline{74.99} & \underline{74.23} & \textbf{72.93} & \textbf{71.80} & \underline{72.15} & \textbf{74.65} & \textbf{70.64} & \textbf{70.65} & \textbf{71.75} & \textbf{74.33} & \textbf{71.76} & \textbf{72.05} & \textbf{73.04}\\
    \cmidrule{1-15}
    \multirow{9}{*}{\makecell[l]{Mini-\\ImageNet}} 
    & Nat. Tra. & \textbf{76.37} & 11.64 & \phantom{0}0.03 & \phantom{0}0.01 & \phantom{0}0.00 & 17.63 & \phantom{0}1.69 & \phantom{0}1.18 & \phantom{0}0.44 & 17.08 & \phantom{0}4.67 & \phantom{0}2.93 & \phantom{0}0.88\\
    &Adv. Tra. & 58.39 & 61.19 & 61.86 & 60.93 & 60.91 & 56.13 & 56.28 & 55.85 & 55.59 & 57.05 & 57.28 & 57.10 & 56.99\\
    &TVM & 66.21 & 64.78 & 66.67 & 64.07 & 64.70 & 59.29 & 61.28 & 58.97 & 59.45 & 62.88 & 64.00 & 62.63 & 63.33\\
    &Feat. Den. & 59.00 & 58.50 & 58.62 & 56.02 & 55.77 & 53.01 & 53.14 & 50.79 & 50.38 & 57.77 & 57.81 & 57.10 & 57.41\\
    &Pix. Def. & 67.28 & 65.17 & 67.28 & 65.43 & 65.14 & 59.74 & 62.01 & 59.93 & 59.76 & 64.54 & 65.40 & 64.40 & 64.48\\
    &Mix. Inf. & 73.81 & 68.50 & 70.00 & 69.23 & 69.26 & 63.41 & 64.74 & \underline{63.75} & 64.08 & 67.67 & \underline{69.17} & 68.58 & 69.09\\
    &HGD & 72.41 & \underline{72.19} & 69.68 & 67.12 & 69.78 & \underline{66.79} & 64.18 & 62.08 & \underline{64.43} & \underline{70.21} & 67.65 & \underline{68.63} & 68.66\\
    &LDT & 73.76 & 70.56 & \underline{71.19} & \underline{69.45} & \textbf{70.69} & 63.25 & \underline{65.14} & 62.21 & 63.68 & 69.25 & 67.32 & 66.08 & \underline{69.39}\\
    &MAPE & \underline{74.86} & \textbf{72.64} & \textbf{71.45} & \textbf{70.14} & \underline{70.16} & \textbf{69.13} & \textbf{70.28} & \textbf{68.80} & \textbf{69.92} & \textbf{75.29} & \textbf{71.32} & \textbf{71.80} & \textbf{72.58}\\
    \bottomrule
  \end{tabularx}
  }
\end{table*}

\begin{table}[t]
    \caption{Classification accuracy rates (\%) in defending against unknown types of adversarial attacks on Mini-ImageNet (\textit{higher is better}). ViT-S/16 and ResNet-34 serve as the substitute model and target model, respectively. For each attack, we show the most successful defense with bold and the second one with underline.}\label{Tab3}
  \centering
  \newcolumntype{C}{>{\centering\arraybackslash}X}
  \newcolumntype{L}[1]{>{}p{#1}}
  \scalebox{0.88}{
  \begin{tabularx}{0.5\linewidth}{L{1.6cm}|CCCCC}
    \toprule
    Defenses & Clean & FGSM & BIM & UPGD & VNIM\\
    \cmidrule{1-6}
    Nat. Tra. & \textbf{76.37} & 42.49 & 39.69 & 34.06 & 26.77\\
    Adv. Tra. & 58.39 & 56.71 & 57.07 & 56.59 & 56.15\\
    TVM & 66.22 & 60.61 & 62.90 & 61.12 & 59.33\\
    Feat. Den. & 59.00 & 54.13 & 56.23 & 54.63 & 53.33\\
    Pix. Def. & 67.28 & 61.29 & \underline{63.50} & 62.08 & 59.53\\
    Mix. Inf. & 73.81 & \underline{61.89} & 62.52 & 61.89 & 61.04\\
    HGD & 72.41 & 60.73 & 62.25 & 62.31 & \underline{61.20}\\
    LDT & 73.76 & 61.76 & 61.97 & \underline{62.65} & 60.27\\
    MAPE & \underline{74.86} & \textbf{64.58} & \textbf{65.43} & \textbf{65.88} & \textbf{64.26}\\
    \bottomrule
  \end{tabularx}
  }
\end{table}

We evaluate the effectiveness of various defense methods against unknown types of adversarial attacks. These attack methods are not used in the training process of the defense methods. They include the one-step method FGSM, the multi-step method BIM, as well as advanced transferable attack methods such as Ultimate PGD (UPGD) and VNIM. The adversarial perturbation budget is $L_{\infty}=8/255$, and the number of steps is set to 50 for iterative attack methods. ResNet-34 serves as the target model under attack, while it, along with ShuffleNet-V2-2$\times$ and ResNet-V2-50, acts as substitute models for launching the attacks. The detailed evaluation results on CIFAR-10, CIFAR-100, and Mini-ImageNet are presented in Table~\ref{Tab2}. For transformer-based models, the selected substitute model is ViT-S/16. The detailed evaluation results on Mini-ImageNet are shown in Table~\ref{Tab3}. ShuffleNet-V2-2$\times$, ResNet-V2-50, and ViT-S/16 are not used in the training process of any defense methods.

In Table~\ref{Tab2}, it can be found that regardless of the defense strategy employed, there will be a reduction in the target model's classification accuracy on clean examples. In comparison, MAPE shows the least degradation of clean classification accuracy. When faced with unknown types of adversarial attacks, MAPE consistently demonstrates superior defensive performance compared to other defense methods, always exhibiting optimal performance against each type of attack. Its average defense performance on CIFAR-10, CIFAR-100, and Mini-ImageNet is 95.03\%, 72.69\%, and 71.41\%, respectively—1.14\%, 2.01\%, and 3.49\% higher than LDT. Furthermore, it can be observed that compared to defending against known substitute models (ResNet-34), all defense methods except for MAPE exhibit a significant performance decline when defending against unforeseen substitute models (ShuffleNet-V2-2$\times$ and ResNet-V2-50). Taking the Mini-ImageNet dataset as an example, the average classification accuracies of HGD, LDT, and MAPE when defending against attacks from ResNet-34 are relatively  similar, at 69.69\%, 70.47\%, and 71.10\%, respectively. However, when defending against attacks from ShuffleNet-V2-2$\times$, their average classification accuracies drop to 64.37\%, 63.57\%, and 69.53\%, representing declines of 5.32\%, 6.9\%, and 1.57\%, respectively. This indicates that, compared to other defense methods, MAPE exhibits higher generalization capabilities, effectively defending against transferable adversarial attacks from unforeseen substitute models.

In Table~\ref{Tab3}, the substitute model ViT-S/16 is based on a transformer architecture, while the target model ResNet-34 relies on convolutional structures. This results in substantial structural differences between them, leading to significant distinctions in their classification boundaries on Mini-ImageNet. Consequently, the effectiveness of adversarial attacks on ViT-S/16 cannot be readily transferred to ResNet-34. Specifically, the natural accuracy drop after experiencing adversarial attacks in Table~\ref{Tab3} is not as pronounced as that in Table~\ref{Tab2}. Similarly, because all defense methods were trained using CNNs as hypothetical substitute models, the effectiveness of defending against adversarial attacks from ViT-S/16 is not as strong as defending against attacks from CNNs. This leads to a curious phenomenon as shown in Table~\ref{Tab3}: the adversarial attacks from ViT-S/16 are not very strong, yet the defensive effectiveness against them is also not very high. Nonetheless, MAPE still demonstrates the strongest defensive capabilities compared to other methods.

\subsection{Defending Against Integrated Transferable Attacks}
\label{Sec44}

We evaluated the effectiveness of different defense methods against integrated adversarial attacks. The integrated adversarial attack methods include LGV \cite{Gub22}, TAIG \cite{Hua22}, and AdaEA \cite{Che23}. The basic attack method, adversarial perturbation budget, and number of steps are set to BIM, $L_{\infty}=8/255$, and 50, respectively. For LGV, the substitute model is ShuffleNet-V2-2$\times$ and the number of weight sets is 10. For TAIG, the substitute model is also ShuffleNet-V2-2$\times$, and the example augmentation factor is 20. For AdaEA, the substitute model ensemble consists of ShuffleNet-V2-2$\times$, ResNet-V2-50, and VGG-19. For all defense methods, these substitute models have never been encountered. The detailed evaluation results on Mini-ImageNet are shown in Table~\ref{Tab4}. It is evident that when confronted with ensemble attacks involving multiple gradients or models, MAPE continues to exhibit the highest natural accuracy and defensive performance compared to other defense methods.

\begin{table}[t]
    \caption{Classification accuracy rates (\%) in defending against integrated adversarial attacks on Mini-ImageNet (\textit{higher is better}). ResNet-34 serve as the target model under attack. For each attack, we show the most successful defense with bold and the second one with underline.}\label{Tab4}
  \centering
  \newcolumntype{C}{>{\centering\arraybackslash}X}
  \newcolumntype{L}[1]{>{}p{#1}}
  \scalebox{0.88}{
  \begin{tabularx}{0.5\linewidth}{L{1.6cm}|CCCC}
    \toprule
    Defenses & Clean & LGV & TAIG & AdaEA\\
    \cmidrule{1-5}
    Nat. Tra. & \textbf{76.37} & \phantom{0}7.03 & \phantom{0}0.37 & \phantom{0}0.98\\
    Adv. Tra. & 58.39 & 57.14 & 55.25 & 56.22\\
    TVM & 66.13 & 62.78 & 58.99 & 59.98\\
    Feat. Den. & 59.00 & 55.52 & 50.18 & 53.66\\
    Pix. Def. & 67.28 & 64.28 & 59.68 & 60.79\\
    Mix. Inf. & 73.81 & 66.08 & \underline{64.23} & 61.73\\
    HGD & 72.41 & 66.72 & 62.58 & 60.67\\
    LDT & 73.76 & \underline{67.40} & 64.10 & \underline{63.41}\\
    MAPE & \underline{74.86} & \textbf{69.43} & \textbf{69.12} & \textbf{69.78}\\
    \bottomrule
  \end{tabularx}
  }
\end{table}

\begin{table*}[t]
    \caption{Classification accuracy rates (\%) in defending against strong substitute model attacks and adaptive attacks on Mini-ImageNet (\textit{higher is better}). For each attack, we show the most successful defense with bold and the second one with underline.}\label{Tab5}
  \centering
  \newcolumntype{C}{>{}X}
  \newcolumntype{L}[1]{>{\centering\arraybackslash}p{#1}}
  \scalebox{0.9}{
  \begin{tabularx}{\textwidth}{C|L{0.89cm}|*{5}{L{0.89cm}}|*{5}{L{0.89cm}}}
    \toprule
    \multirow{2}{*}{Defenses} & \multirow{2}{*}{Clean} & \multicolumn{5}{c|}{Strong Substitute Model Attacks} & \multicolumn{5}{c}{Adaptive Attacks}\\[0.4ex]
    && FGSM & BIM & UPGD & VMIM & VNIM & FGSM & BIM & UPGD & VMIM & VNIM\\
    \cmidrule{1-12}
    Nat. Tra. & \textbf{76.37} & 13.40 & \phantom{0}3.39 & \phantom{0}2.03 & \phantom{0}1.19 & \phantom{0}1.07 & 11.69 & \phantom{0}0.03 & \phantom{0}0.03 & \phantom{0}0.01 & \phantom{0}0.01\\
    Adv. Tra. & 58.39 & 32.27 & 29.22 & 29.42 & 29.87 & 29.04 & 21.37 & \textbf{15.42} & \textbf{16.26} & \textbf{16.14} & \textbf{16.33}\\
    TVM & 66.17 & 50.85 & 45.62 & 46.78 & 40.49 & 39.48 & 18.84 & \underline{\phantom{0}4.49} & \underline{\phantom{0}5.16} & \underline{\phantom{0}6.77} & \underline{\phantom{0}7.38}\\
    Feat. Den. & 59.00 & 21.45 & 12.17 & 11.52 & 11.39 & 10.57 & \phantom{0}6.75 & \phantom{0}0.23 & \phantom{0}0.31 & \phantom{0}0.33 & \phantom{0}0.31\\
    Pix. Def. & 67.28 & 41.04 & 35.88 & 34.71 & 34.37 & 33.67 & 15.63 & \phantom{0}0.57 & \phantom{0}0.73 & \phantom{0}0.76 & \phantom{0}0.78\\
    Mix. Inf. & 73.81 & 42.73 & 37.70 & 35.48 & 32.45 & 29.98 & \textbf{29.46} & \phantom{0}0.00 & \phantom{0}0.00 & \phantom{0}0.01 & \phantom{0}0.00\\
    HGD & 72.41 & 51.72 & 40.81 & 38.77 & 35.78 & 33.38 & 13.93 & \phantom{0}0.05 & \phantom{0}0.06 & \phantom{0}0.46 & \phantom{0}0.52\\
    LDT & 73.76 & \underline{55.32} & \underline{46.77} & \underline{46.87} & \underline{44.80} & \underline{41.58} & 19.97 & \phantom{0}0.06 & \phantom{0}0.10 & \phantom{0}0.79 & \phantom{0}1.48\\
    MAPE & \underline{74.86} & \textbf{63.62} & \textbf{54.42} & \textbf{55.89} & \textbf{53.53} & \textbf{51.29} & \underline{21.98} & \phantom{0}0.11 & \phantom{0}0.12 & \phantom{0}1.68 & \phantom{0}2.34\\
    \bottomrule
  \end{tabularx}
  }
\end{table*}

\subsection{Defending Against Strong Substitute Model Attacks and Adaptive Attacks}
\label{Sec45}

In this section, we consider the defensive effect of the proposed method in gray-box and white-box attack environments. In a gray-box attack environment, the attackers employ strong substitute model attacks. In a white-box attack environment, attackers use adaptive attacks. The conditions for carrying out white-box attacks are more rigorous than those for gray-box attacks. A detailed overview is provided in Subsection~\ref{Sec41}.

We evaluated the effectiveness of different defense methods against strong substitute model attacks and adaptive attacks. The base target model under attack is ResNet-34. The adversarial attack methods include FGSM, BIM, UPGD, VMIM, and VNIM. The adversarial perturbation budget is $L_{\infty}=8/255$, and the number of steps is set to 50 for iterative attack methods. For defense methods based on obfuscated gradients or random transformation, we respectively employ backward pass differentiable approximation (BPDA) \cite{AthObf18} and expectation over transformation (EOT) \cite{AthObf18} for gradient correction. The detailed evaluation results on Mini-ImageNet are presented in Table ~\ref{Tab5}.

From Table~\ref{Tab5}, it is evident that in strong substitute model attacks under gray-box environments, MAPE demonstrates superior defense effectiveness compared to other methods. This is because, during the training process of MAPE, multiple classification models are introduced to aid in training, alongside random adjustments to adversarial example generation methods and perturbation budgets. Consequently, among all defense approaches, MAPE exhibits the largest sample space and parameter space, significantly reducing the similarity in parameter weights between strong substitute models and MAPE. However, in adaptive attacks under white-box environments, where all model weights and defense strategies are exposed, attackers can develop precise attack strategies, resulting in a significant decrease in the defensive capabilities of all methods. Although adversarial training and TVM can defend against a small number of adversarial examples, their natural accuracy and black-box accuracy are far inferior to those of MAPE. 

Failure of defense against adaptive attacks is not a unique issue of the proposed method but rather a common problem associated with black-box defense methods. Such failures can also be found in the original papers on HGD \cite{Lia18} and LDT \cite{Li23}. Although this paper primarily focuses on the domain of black-box defenses, integrating the proposed method with several advanced adversarial training techniques can still enhance its applicability in the field of white-box defenses. For instance, employing classical TRADES \cite{Zha_The19} and MART \cite{Wan20} loss functions in the adversarial training process of the proposed method can optimize the classification boundaries of the model. Utilizing data generated by the latest elucidating diffusion model (EDM) as training data \cite{Wan_Bet23} can simultaneously improve the performance of the proposed method in both black-box and white-box defenses. Additionally, a specific scaling law \cite{Bar24} allows for more rational allocation of resources such as model and dataset sizes, thereby maximizing adversarial robustness given a fixed computational capacity. Furthermore, combining the fast adversarial training method known as FGSM-PCO \cite{Wan_Pre24} can reduce the computational costs associated with the adversarial training process. In summary, integrating adversarial training methods aids in improving the white-box robustness of MAPE, potentially broadening the applicability of the proposed method. This will be our primary research direction moving forward.

\begin{table*}[t]
    \caption{Classification accuracy rates (\%) of different methods for different target models in defending against unknown types of adversarial attacks on CIFAR-10, CIFAR-100 and Mini-ImageNet (\textit{higher is better}). ShuffleNet-V2-2$\times$ and ResNet-V2-50 serve as substitute models for generating adversarial examples, VGG-19 and ViT-S/16 serve as other target models not included in the framework. For each attack, we show the most successful defense with bold.}\label{Tab6}
  \centering
  \newcolumntype{C}{>{}X}
  \newcolumntype{L}[1]{>{\centering\arraybackslash}p{#1}}
  \scalebox{0.88}{
  \begin{tabularx}{\textwidth}{CC|L{0.76cm}|*{5}{L{0.76cm}}|*{5}{L{0.76cm}}}
    \toprule
    \multirow{2}{*}{Dataset} & \multirow{2}{*}{Defenses} & \multirow{2}{*}{Clean} & \multicolumn{5}{c|}{ShuffleNet-V2-2$\times$} & \multicolumn{5}{c}{ResNet-V2-50}\\[0.4ex]
    &&& FGSM & BIM & UPGD & VMIM & VNIM & FGSM & BIM & UPGD & VMIM & VNIM\\
    \cmidrule{1-13}
    \multicolumn{13}{c}{VGG-19 serves as the target model under attack.}\\
    \cmidrule{1-13}
    \multirow{4}{*}{CIFAR-10}
    &Nat. Tra. & \textbf{94.69} & 51.11 & 18.55 & 13.14 & \phantom{0}5.67 & \phantom{0}5.96 & 48.31 & 22.41 & 15.90 & \phantom{0}7.54 & \phantom{0}7.21\\
    &HGD & 90.46 & 89.75 & 90.19 & 89.79 & 90.06 & 90.22 & 92.05 & 91.35 & 91.13 & 91.53 & 91.64\\
    &LDT & 93.03 & 91.32 & 90.67 & 90.56 & 90.77 & 90.74 & 92.21 & 91.69 & 91.87 & 92.56 & 92.69\\
    &MAPE & 94.31 & \textbf{94.10} & \textbf{93.74} & \textbf{94.18} & \textbf{94.11} & \textbf{93.89} & \textbf{93.94} & \textbf{94.02} & \textbf{94.13} & \textbf{94.29} & \textbf{93.84}\\
    \cmidrule{1-13}
    \multirow{4}{*}{CIFAR-100}   
    &Nat. Tra. & \textbf{75.16} & 26.77 & 14.22 & 10.57 & \phantom{0}6.93 & \phantom{0}6.75 & 30.13 & 23.97 & 19.24 & 11.26 & 10.45\\
    &HGD & 67.38 & 63.53 & 61.92 & 60.73 & 62.16 & 62.59 & 65.20 & 62.47 & 60.82 & 62.71 & 63.01\\
    &LDT & 71.43 & 67.48 & 62.15 & 60.70 & 63.14 & 62.91 & 69.48 & 63.11 & 63.50 & 64.73 & 65.11\\
    &MAPE & 74.48 & \textbf{72.82} & \textbf{69.08} & \textbf{69.81} & \textbf{70.31} & \textbf{70.28} & \textbf{73.89} & \textbf{69.71} & \textbf{70.92} & \textbf{72.24} & \textbf{72.46}\\
    \cmidrule{1-13}
    \multirow{4}{*}{\makecell[l]{Mini-\\ImageNet}} 
    &Nat. Tra. & \textbf{70.33} & 21.96 & 11.10 & \phantom{0}8.41 & \phantom{0}4.02 & \phantom{0}3.65 & 22.49 & 29.37 & 24.43 & 12.56 & 10.63\\
    &HGD & 62.83 & 57.73 & 58.88 & 56.36 & 57.37 & 58.09 & 61.93 & 61.00 & 58.83 & 60.92 & 61.32\\
    &LDT & 65.82 & 63.97 & 59.64 & 58.25 & 59.58 & 59.88 & 66.71 & 62.91 & 63.32 & 64.93 & 65.11\\
    &MAPE & 69.27 & \textbf{64.59} & \textbf{65.19} & \textbf{64.73} & \textbf{65.11} & \textbf{64.99} & \textbf{68.44} & \textbf{66.47} & \textbf{66.68} & \textbf{67.52} & \textbf{67.15}\\
    \cmidrule{1-13}
    \multicolumn{13}{c}{ViT-S/16 serves as the target model under attack.}\\
    \cmidrule{1-13}
    \multirow{4}{*}{CIFAR-10}
    &Nat. Tra. & \textbf{98.67} & 85.30 & 82.77 & 78.26 & 67.31 & 68.07 & 88.01 & 85.03 & 81.21 & 73.14 & 73.58\\
    &HGD & 95.20 & 94.07 & 93.76 & 93.52 & 93.48 & 93.43 & 95.14 & 94.30 & 94.15 & 94.71 & 94.32\\
    &LDT & 97.22 & 96.04 & 95.60 & 95.41 & 95.42 & 95.53 & 96.89 & 95.80 & 95.79 & 96.04 & 96.24\\
    &MAPE & 98.53 & \textbf{97.06} & \textbf{96.19} & \textbf{96.27} & \textbf{96.35} & \textbf{96.55} & \textbf{97.47} & \textbf{96.87} & \textbf{96.72} & \textbf{96.85} & \textbf{96.74}\\
    \cmidrule{1-13}
    \multirow{4}{*}{CIFAR-100}   
    &Nat. Tra. & \textbf{90.32} & 64.03 & 60.51 & 55.15 & 49.08 & 49.35 & 66.84 & 66.59 & 61.31 & 52.70 & 53.51\\
    &HGD & 78.94 & 74.99 & 73.97 & 72.57 & 73.43 & 73.47 & 74.73 & 72.64 & 71.34 & 72.08 & 72.96\\
    &LDT & 82.46 & 80.01 & 74.75 & 74.23 & 75.76 & 76.26 & 79.77 & 73.77 & 74.39 & 76.10 & 76.12\\
    &MAPE & 89.80 & \textbf{85.94} & \textbf{82.20} & \textbf{82.47} & \textbf{83.34} & \textbf{83.63} & \textbf{87.76} & \textbf{83.13} & \textbf{83.82} & \textbf{85.75} & \textbf{85.53}\\
    \cmidrule{1-13}
    \multirow{4}{*}{\makecell[l]{Mini-\\ImageNet}} 
    &Nat. Tra. & \textbf{90.53} & 68.77 & 69.18 & 63.00 & 52.30 & 50.86 & 73.45 & 76.28 & 71.03 & 59.88 & 60.38\\
    &HGD & 84.36 & 80.04 & 79.74 & 78.03 & 79.57 & 79.89 & 82.16 & 81.55 & 80.11 & 81.06 & 81.43\\
    &LDT & 87.19 & 84.97 & 83.32 & 82.42 & 83.39 & 83.68 & 86.80 & 84.33 & 84.18 & 85.30 & 85.58\\
    &MAPE & 89.38 & \textbf{86.45} & \textbf{85.73} & \textbf{85.47} & \textbf{86.04} & \textbf{86.07} & \textbf{88.09} & \textbf{86.64} & \textbf{86.68} & \textbf{87.45} & \textbf{87.43}\\
    \bottomrule
  \end{tabularx}
  }
\end{table*}

\section{Further Evaluations}
\label{Sec5}

\subsection{Cross-Model Defense}

We evaluated the transferability of the defensive capabilities across different methods, i.e., their defensive effects on different target models. The adversarial attack methods include FGSM, BIM, UPGD, VMIM, and VNIM. The adversarial perturbation budget is $L_{\infty}=8/255$, and the number of steps is set to 50 for iterative attack methods. The selected substitute models are ShuffleNet-V2-2$\times$ and ResNet-V2-50. We only choose to compare HGD and LDT with MAPE because their defense and target models can be separated, and the target model is independently trained (not jointly trained with the defense model). Therefore, when the original target model is replaced with different target models VGG-19 and ViT-S/16, they can still function properly. At this point, their defensive effects depend on the robustness and generalization of the defense model. The detailed evaluation results on CIFAR-10, CIFAR-100, and Mini-ImageNet are presented in Table~\ref{Tab6}.

From Table~\ref{Tab6}, it can be observed that the natural accuracy of MAPE closely aligns with the natural accuracy of the target model. This suggests that MAPE rarely leads to misclassification of input examples by the target model. When compared to HGD and LDT, MAPE achieves the best defensive performance under each type of attack, whether assisting the CNN model VGG-19 or the transformer model ViT-S/16. This indicates that MAPE is capable of training a defense model with strong generalization capabilities. This defense model is independent of the target model, and its performance is not affected by it. After training, it can provide adversarial defense for different target models without the need for retraining for each specific target model.

\subsection{Ablation Study}

\begin{figure}[t]
\centering
\includegraphics[width=0.45\linewidth]{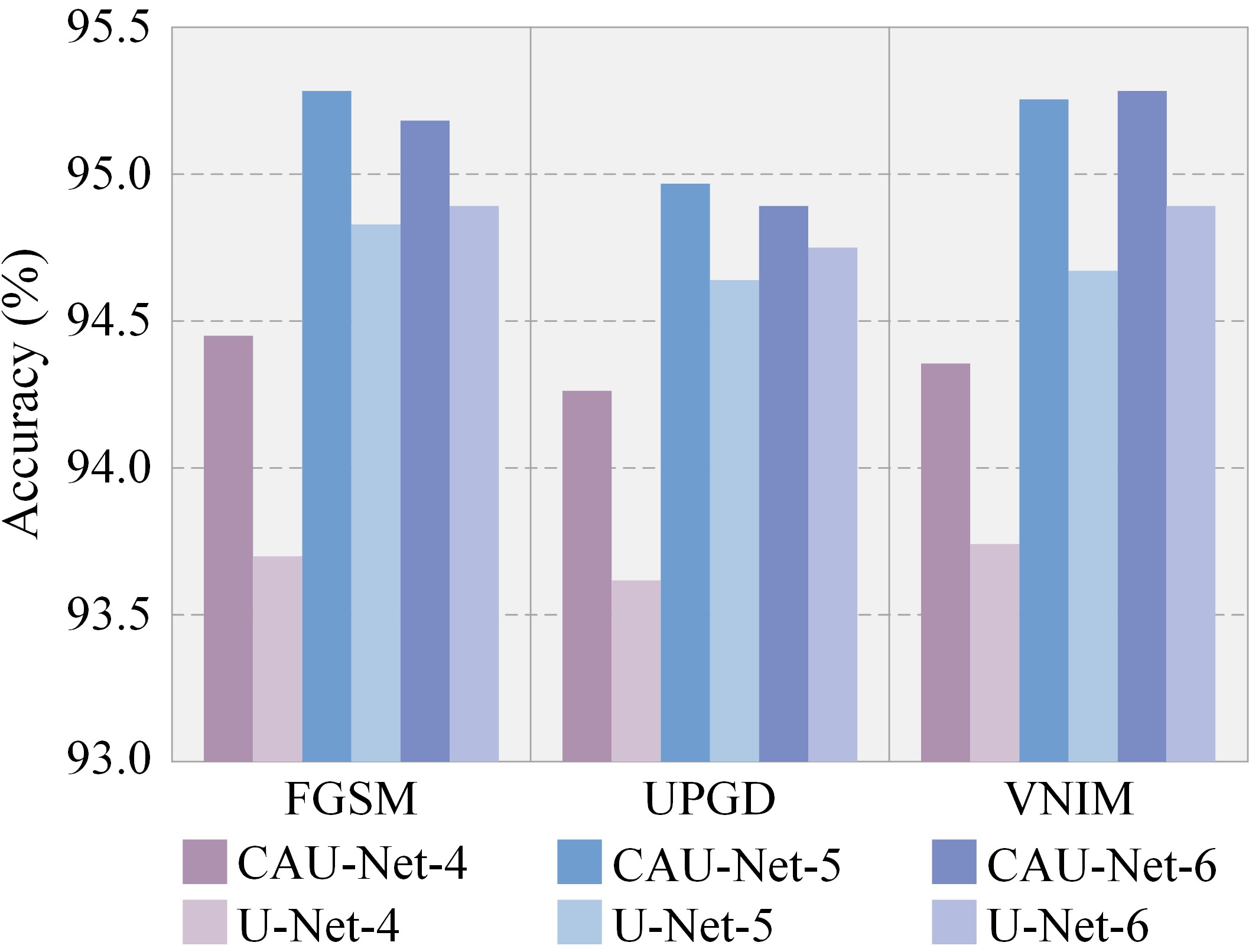}
\caption{Classification accuracy rates (\%) of MAPEs with different defense models in defending against unknown types of adversarial attacks on CIFAR-10 (\textit{higher is better}). ShuffleNet-V2-2$\times$ and ResNet-34 serve as the substitute model and target model, respectively. CA is channel-attention mechanism and numerical suffixes is the quantity of submodules.}
\label{Fig5}
\end{figure}

Based on whether channel-attention mechanism layers have been added, we divide the defense models into CAU-Net and U-Net. The former incorporates channel-attention mechanism layers, while the latter does not. Subsequently, we assign 4, 5, and 6 submodules to each, resulting in a total of 6 defense models participating in the ablation study. The defense model utilized in this paper is CAU-Net-5, which features channel-attention mechanism layers and 5 submodules. Each defense model is trained in a uniform manner and then subjected to adversarial attacks including FGSM, UPGD, and VNIM. The adversarial perturbation budget is $L_{\infty}=8/255$, and the number of steps is set to 50. ShuffleNet-V2-2$\times$ serves as the substitute model for generating adversarial examples, and ResNet-34 serves as the target model under attack. 

Figure~\ref{Fig5} presents the ablation study of the defense models discussed above on CIFAR-10. It can be observed that when the number of submodules (network depth) is the same, CAU-Net with channel-attention mechanism layers exhibits stronger robustness and better performance in defending against adversarial attacks compared to U-Net. The relatively shallow network depth of CAU-Net-4 results in a diminished fitting capability and a lower defense effectiveness. Conversely, excessive network depth in CAU-Net-6 may lead to overfitting issues, resulting in a decline in its defense performance. Therefore, among the three models, CAU-Net-5 exhibits a more suitable fitting capability and the strongest overall defense performance. Additionally, with only 1.66M parameters, CAU-Net-4 comprises merely 24.85\% of CAU-Net-5's parameter count (6.68M). Given a modest compromise in defense performance, CAU-Net-4 can be employed for adversarial defense in lightweight classification models.

\subsection{Robustness Against Perturbation Budgets}

\begin{figure}[t]
\centering
\includegraphics[width=0.45\linewidth]{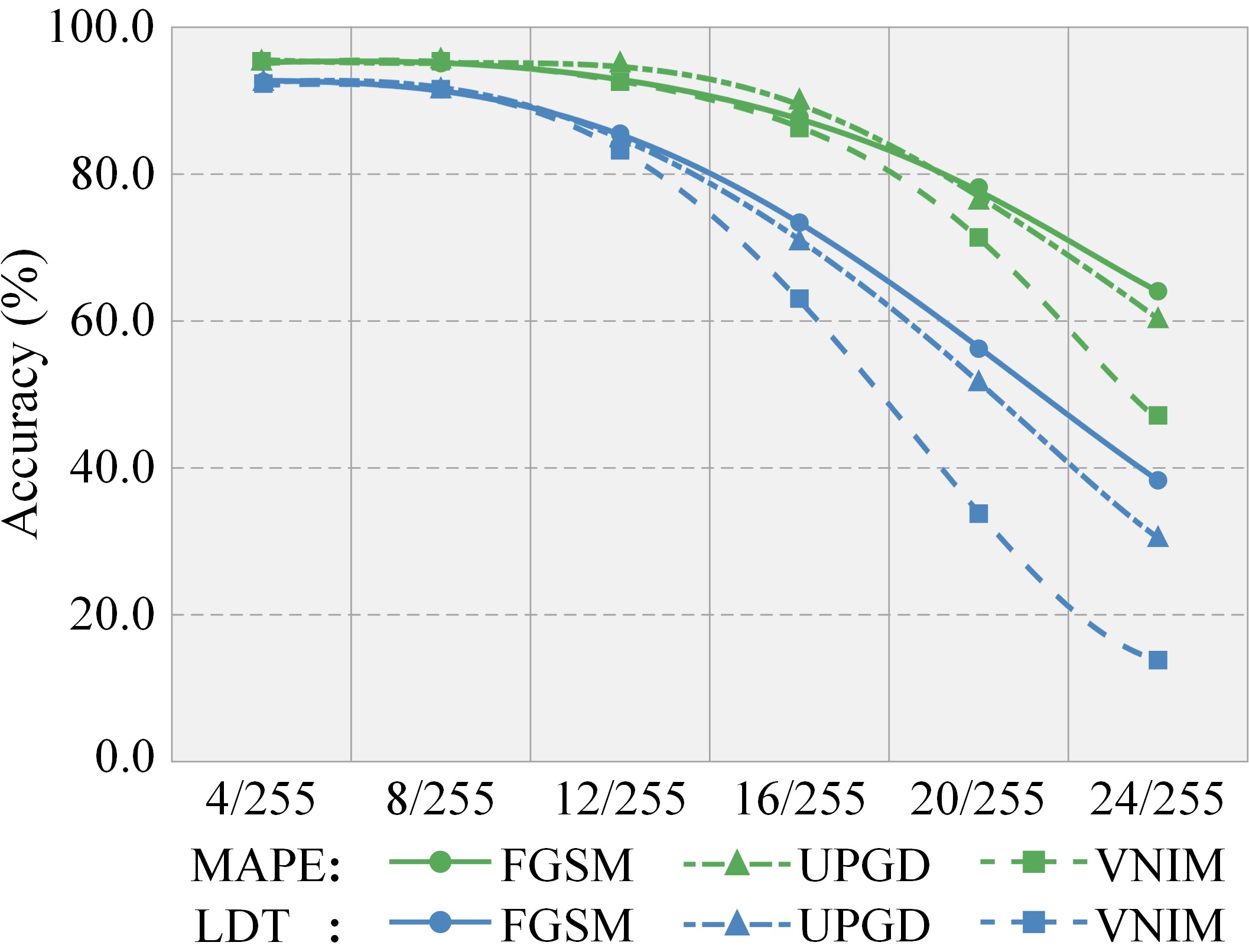}
\caption{Classification accuracy rates (\%) of MAPE and LDT in defending against unknown types of adversarial attacks with different perturbation budgets on CIFAR-10 (\textit{higher is better}). ShuffleNet-V2-2$\times$ and ResNet-34 serve as the substitute model and target model, respectively.}
\label{Fig6}
\end{figure}

We set the $L_\infty$ norm perturbation budget within the range of $[4/255, 24/255]$ and employ FGSM, UPGD, and VNIM to attack the target model, in order to evaluate the robustness of MAPE against perturbation budgets. The detailed evaluation results are shown in Figure~\ref{Fig6}. It can be observed that when the adversarial perturbation budget is less than 12/255, LDT is able to maintain a defense effectiveness of over 80\%, but as the perturbation budget increases, the defense effectiveness sharply decreases. In comparison, it is only when the adversarial perturbation budget surpasses 16/255 that MAPE's defense effectiveness shows a significant decrease. Furthermore, at an adversarial perturbation budget of 4/255, MAPE exhibits an average defense effectiveness 3.1\% higher than LDT. However, at a perturbation budget of 24/255, MAPE's average defense effectiveness surpasses LDT by a remarkable 105.9\%. This indicates that the defense based on MAPE exhibits strong robustness against high adversarial perturbation budgets and powerful adversarial attacks.

\subsection{Defense Costs}

\begin{figure}[t]
  \centering
  \includegraphics[width=0.91\linewidth]{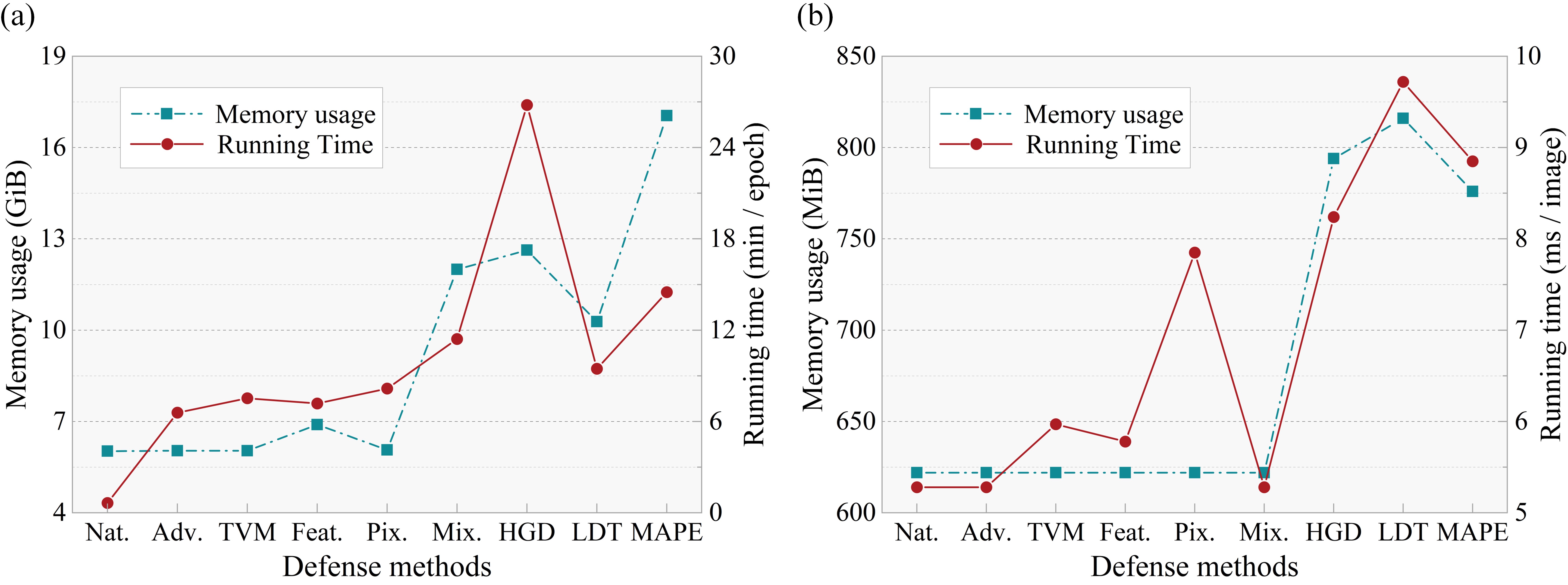}
  \caption{Comparison of different defense methods in terms of (a) training cost and (b) evaluation cost. The comparison metrics include the memory usage and the running time. The target model and dataset used are ResNet-34 and Mini-ImageNet, respectively.}
  \label{Fig7}
\end{figure}

We compared the training and evaluation costs of different defense methods on Mini-ImageNet, as shown in Figure~\ref{Fig7}. The evaluation cost refers to the actual operational cost. The comparison metrics include memory usage and running time. All defense methods employed a pre-trained ResNet-34 as the target model. When measuring the training cost, the batch size of the dataset was set to 128. For evaluating the cost, defense methods processed a single image at a time, running a total of 10,000 iterations, with the average taken as the processing cost for one image.

During the training process, methods incorporating deep defense models, such as HGD, LDT, and proposed MAPE, require higher training costs compared to others. HGD employs the target model to predict both complete clean examples and adversarial examples separately, so it has the longest running time. In contrast, LDT and MAPE only require predictions for mixed examples composed of clean and adversarial examples, so their running times are considerably shorter than those of HGD. MAPE appears to consume a substantial amount of memory when contrasted with HGD and LDT. However, due to the memory reuse mechanism in PyTorch, the memory usage of MAPE is not the sum of the memory usage of the pre-trained models, but rather their upper bound. This implies that certain heavyweight pre-trained models, such as PyramidNet and WideResNet, are responsible for the increased memory usage. In practical applications, if memory is constrained, these heavier models can be replaced with lighter alternatives. 

During the evaluation process, HGD, LDT, and MAPE still incur higher defense costs compared to other methods. However, due to the utilization of a more efficient CAU-Net in MAPE, its memory usage is 2.3\% (18 MiB) lower than that of HGD and 4.9\% (40 MiB) lower than that of LDT. In terms of runtime for a single image, MAPE takes 9.0\% (0.87 ms) less time than LDT, while it takes 7.4\% (0.61 ms) more time than HGD. Furthermore, the model parameter counts for HGD, LDT, and MAPE are 32.36M, 33.44M, and 28.01M, respectively, while the parameter counts for other methods are approximately 21.3M. The model parameter count of MAPE is 13.4\% (4.35M) lower than that of HGD and 16.3\% (5.43M) lower than that of LDT.

In summary, compared to HGD and LDT, MAPE exhibits the lowest memory usage during evaluation while demonstrating the highest memory usage during training. When memory is constrained, the latter issue can be addressed by substituting lighter pre-trained models. In terms of running time, MAPE consistently maintains an intermediate level of performance. Additionally, MAPE has the lowest model parameter count. Combining these findings with previous defense experiment results, it is evident that MAPE not only possesses the strongest defense performance but also incurs a defense cost comparable to other methods of similar type.

\section{Conclusion}

In this paper, our approach is to deploy a defense model external to the target model to extract and eliminate the adversarial perturbations from input examples. The optimization objective focuses on enhancing the robustness and generalization of the used defense model to effectively defend against a variety of unknown types of adversarial attacks. To achieve this goal, we propose a deep learning defense known as MAPE, which is primarily composed of SAPE and PPSA. SAPE utilizes CAU-Net as its defense model, training it to eliminate adversarial perturbations by using adversarial examples generated from a pre-trained model. Meanwhile, PPSA integrates model difference probability and negative momentum probability to strategically schedule multiple pre-trained models, maximizing the differences among these models during adjacent training cycles, thus enhancing the diversity of the generated adversarial examples. The evaluation results demonstrate that MAPE exhibits strong robustness and can effectively defend against various types of adversarial attacks in a black-box environment. In future work, potential extensions include strengthening its defense capabilities against strong substitute model attacks and adaptive attacks, as well as applying it to adversarial defense in object detection, image segmentation, and other visual tasks.

\bibliographystyle{cas-model2-names}

\bibliography{complete_cas-refs}

\end{document}